\documentclass[twocolumn,letterpaper]{article}

\PassOptionsToPackage{table}{xcolor}
\usepackage[numbers,compress]{natbib}
\usepackage{paper-theme}
\usepackage{fontawesome5}

\definecolor{tableblue}{HTML}{E8F1F7}

\setrunningtitle{LeVLJEPA}
\setsectionnavigation
  {1: Intro}
  {2: Background}
  {3: Architecture}
  {4: Experiments}
  {5: Global Features}
  {6: Dense Features}
  {7: Discussion}

\papertitle{LeVLJEPA: End-to-End Vision-Language Pretraining\\ Without Negatives}
\paperauthors{%
  \textbf{Lukas Kuhn}\textsuperscript{1,2}\quad
  \textbf{Giuseppe Serra}\textsuperscript{1}\quad
  \textbf{Randall Balestriero}\textsuperscript{*,4}\quad
  \textbf{Florian Buettner}\textsuperscript{*,1,2,3}}
\paperaffiliations{%
  \textsuperscript{1}German Cancer Research Center\quad
  \textsuperscript{2}German Cancer Consortium\quad
  \textsuperscript{3}Goethe University Frankfurt\quad
  \textsuperscript{4}Brown University}
\papercontribution{\textsuperscript{*}Joint last author.\quad Correspondence: \texttt{lukas.kuhn@dkfz-heidelberg.de}}

\begin{document}

\newcommand{\mstd}[2]{\ensuremath{#1_{\scriptscriptstyle\pm #2}}}
\newcommand{\bmstd}[2]{\ensuremath{\mathbf{#1}_{\scriptscriptstyle\pm #2}}}

\setcounter{topnumber}{2}
\setcounter{totalnumber}{3}
\renewcommand{\topfraction}{0.9}
\renewcommand{\textfraction}{0.1}
\renewcommand{\floatpagefraction}{0.8}
\renewcommand{\dbltopfraction}{0.9}
\renewcommand{\dblfloatpagefraction}{0.8}

\PaperMakeTitle[%
\vspace*{-1.0em}
\begin{center}
  \paperlink{\faGlobe\hspace{0.45em}Website}{https://levljepa.github.io}\hspace{0.45em}
  \paperlink{\faGithub\hspace{0.45em}Code}{https://github.com/mlo-lab/LeVLJEPA}\hspace{0.45em}
  \paperlink{\faRobot\hspace{0.45em}Checkpoints}{https://huggingface.co/lukaskuhndkfz/LeVLJEPA-ViT-B-DataComp-200k}
\end{center}
\begin{paperabstract}

Vision-language pretraining remains dominated by contrastive objectives, whereas vision-only self-supervised learning has largely adopted non-contrastive methods. At the same time, the role of vision-language encoders has shifted: they are increasingly deployed not as zero-shot classifiers but as the frozen visual backbone of vision-language models and dense prediction systems, which consume the full grid of patch tokens rather than a single pooled embedding. We introduce \textbf{LeVLJEPA}, the first fully non-contrastive end-to-end vision-language pretraining method. LeVLJEPA learns through cross-modal prediction with stop-gradient targets and per-modality distributional regularization, without negatives, temperature, momentum encoder, or teacher-student schedule, and trains stably at large scale. We find that the resulting encoder provides markedly stronger dense semantic features for downstream use: as a frozen vision-language-model backbone, LeVLJEPA is the strongest of the evaluated encoders across GQA, VQAv2, and POPE under two distinct language models, and outperforms contrastive baselines on semantic segmentation, while remaining on par on global readouts such as linear probing. These results establish non-contrastive pretraining as an effective means of producing dense semantic vision features.

\end{paperabstract}

]

\section{Introduction}

Self-supervised pretraining for vision has increasingly shifted toward non-contrastive objectives. Methods such as DINO~\citep{caron2021emerging}, I-JEPA~\citep{assran2023self}, and LeJEPA~\citep{balestriero2025lejepa} learn strong visual representations without negative pairs, relying instead on predictive or distributional objectives. Vision-language pretraining has not followed the same trajectory: CLIP~\citep{radford2021learning}, SigLIP~\citep{zhai2023sigmoid}, and their extensions still use contrastive image-text alignment as the default objective for learning from paired image-caption data. This raises a basic question: can end-to-end vision-language pretraining work without contrastive negatives?

This question has practical implications. Vision-language encoders are increasingly reused as visual backbones in systems where the quality of the image representation matters directly, including visual instruction-tuned models such as LLaVA~\citep{liu2023visual}, dense prediction methods such as DenseCLIP~\citep{rao2022denseclip}, and language-conditioned robotics systems such as CLIPort~\citep{shridhar2022cliport} and Voltron~\citep{karamcheti2023language}. In many such settings the pretrained vision encoder serves as a frozen feature extractor, and the downstream system does not consume a single pooled image vector but attends over the full grid of patch tokens. What matters is therefore the quality of these dense, per-token visual features rather than zero-shot classification alone. Yet contrastive vision-language objectives optimize primarily for cross-modal alignment of a single pooled embedding which is well suited to zero-shot retrieval and classification, but is supervising the per-token features only as a byproduct.

We introduce \textbf{LeVLJEPA}, the first fully non-contrastive end-to-end vision-language pretraining method. LeVLJEPA learns image-text alignment through cross-modal prediction: image embeddings predict stop-gradient text embeddings, and text embeddings predict stop-gradient image embeddings through modality-specific predictors. To prevent collapse, each modality is regularized independently with SIGReg~\citep{balestriero2025lejepa}, encouraging isotropic-Gaussian marginal embedding distributions. This gives a simple training recipe with no negatives, temperature parameter, momentum encoder, or teacher-student schedule.

\begin{figure*}[t]
  \centering
  \includegraphics[width=\linewidth]{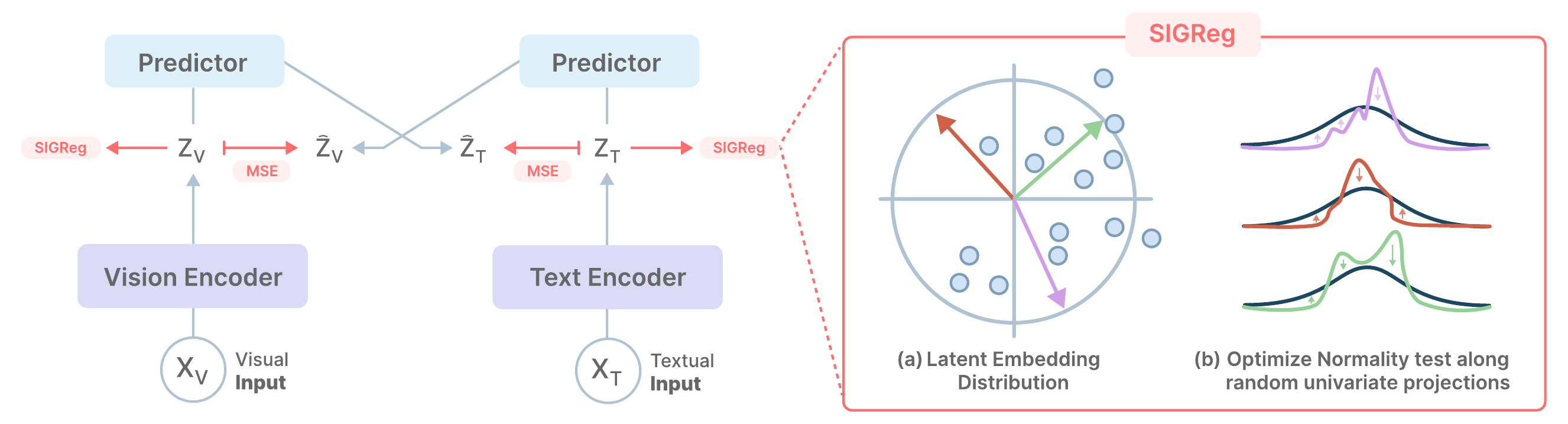}
   \caption{\textbf{LeVLJEPA overview.} Image $X_V$ and text $X_T$ are encoded into  embeddings $Z_V, Z_T$, which are then passed through modality-specific predictors to produce cross-modal predictions $\hat{Z}_V, \hat{Z}_T$. The training objective combines (i) a cross-modal MSE loss between each predictor's output and the stop-gradient target from the other modality, and (ii) SIGReg~\citep{balestriero2025lejepa} applied independently to $Z_V$ and $Z_T$ to keep each modality's marginal embedding distribution close to isotropic Gaussian. SIGReg (right) enforces this distribution by projecting embeddings onto random univariate directions and applying a characteristic-function-based normality test to each projection, avoiding the curse of dimensionality of direct density matching.}
   
   \label{fig:overview}
\end{figure*}

We validate our method at two scales, training on CC12M (12M samples) \citep{changpinyo2021conceptual} and on Datacomp-L (92M samples) ~\citep{gadre2023datacomp}. At the smaller CC12M scale, LeVLJEPA is competitive with CLIP and SigLIP on both zero-shot transfer and linear probing, establishing that the non-contrastive objective matches contrastive pretraining when the alignment signal is the primary axis of evaluation. At the larger Datacomp-L scale, contrastive objectives retain an advantage on zero-shot transfer, while LeVLJEPA remains on par with CLIP and SigLIP under linear probing of the global class representation and outperforms both on background robustness. The objectives diverge most strongly under evaluations that operate on the patch-token sequence rather than a pooled embedding. LeVLJEPA outperforms both contrastive baselines on semantic segmentation, and when employed as the frozen visual backbone of a vision-language model---with only a lightweight bridge trained between the frozen encoder and a frozen language model---it attains the highest downstream accuracy on GQA, VQAv2, and POPE, consistently across two distinct language-model families. These results indicate that the benefit of non-contrastive vision-language pretraining is concentrated in the quality of the dense, per-token features on which downstream systems operate, rather than in the global image-text alignment measured by zero-shot transfer.

The contributions of this work are as follows:
\begin{itemize}
    \item We introduce \textbf{LeVLJEPA}, the first fully non-contrastive end-to-end vision-language pretraining method. The objective combines cross-modal prediction with stop-gradient targets and per-modality distributional regularization, and requires no negative pairs, temperature, momentum encoder, or teacher-student schedule, yielding a simple and stable training procedure.
    \item We establish that our method is competitive with contrastive pretraining across two training scales. At smaller scale training it matches CLIP and SigLIP on both zero-shot transfer and linear probing, while at the scale of Datacomp-L it remains on par under linear probing and trails the contrastive baselines only on zero-shot transfer, the objective these methods optimize directly.
    \item We demonstrate that LeVLJEPA learns stronger dense semantic features for downstream use. Evaluated under protocols that operate on the patch-token sequence rather than a pooled embedding, it outperforms both contrastive baselines on semantic segmentation and, as the frozen visual backbone of a vision-language model, attains the highest accuracy on GQA, VQAv2, and POPE across two distinct language-model families.
\end{itemize}

\section{Background}

\subsection{Notation and Setup}

We consider a paired dataset $\mathcal{D} = \{(x_i, y_i)\}_{i=1}^{N}$ of images $x_i \in \mathcal{X}$ and captions $y_i \in \mathcal{Y}$. An image encoder $f_\theta : \mathcal{X} \to \mathbb{R}^d$ maps images to embeddings $z^v = f_\theta(x)$, and a text encoder $g_\phi : \mathcal{Y} \to \mathbb{R}^d$ maps captions to embeddings $z^t = g_\phi(y)$. A minibatch of size $B$ yields embedding matrices $Z^v, Z^t \in \mathbb{R}^{B \times d}$ with rows $z^v_i, z^t_i$. For contrastive baselines and evaluation, embeddings are $\ell_2$-normalized before computing cosine similarities. For LeVLJEPA, the embeddings regularized by SIGReg are not $\ell_2$-normalized during training, since SIGReg targets an isotropic Gaussian marginal distribution in $\mathbb{R}^d$. When cosine similarities are needed for zero-shot evaluation or retrieval, LeVLJEPA embeddings are normalized only at evaluation time.

\subsection{Contrastive Vision-Language Pretraining}

CLIP~\citep{radford2021learning} trains $f_\theta$ and $g_\phi$ jointly by pulling matched image-caption pairs together and pushing unmatched pairs apart, using a symmetric InfoNCE objective over the batch,
\begin{equation}
\begin{split}
    \mathcal{L}_{\text{InfoNCE}} = -\frac{1}{2B} \sum_{i=1}^{B} \Bigg[ &\log \frac{\exp(z^v_i \cdot z^t_i / \tau)}{\sum_{j=1}^{B} \exp(z^v_i \cdot z^t_j / \tau)} \\
    &+ \log \frac{\exp(z^v_i \cdot z^t_i / \tau)}{\sum_{j=1}^{B} \exp(z^v_j \cdot z^t_i / \tau)} \Bigg],
\end{split}
\end{equation}
where $\tau$ is a learned temperature. The softmax over the batch turns every other sample into a negative, so the quality of the signal scales with $B$: larger batches provide harder and more diverse negatives, and CLIP-style methods are typically trained with batch sizes in the tens of thousands.

SigLIP~\citep{zhai2023sigmoid} replaces the softmax with an independent sigmoid classification over every pair $(i, j)$ in the batch,
\begin{equation}
    \mathcal{L}_{\text{SigLIP}} = -\frac{1}{B} \sum_{i=1}^{B} \sum_{j=1}^{B} \log \sigma\!\left( t_{ij} \left( z^v_i \cdot z^t_j + b \right) \right),
\end{equation}
where $t_{ij} = +1$ if $i = j$ and $-1$ otherwise, and $t, b$ are learnable scalars. Decoupling the pairs removes the global normalization and improves training stability at small and large batch sizes alike, though performance still benefits meaningfully from scaling $B$.

VL-JEPA~\citep{chen2025vljepa} recasts vision-language learning in the language of joint-embedding predictive architectures: rather than aligning two independently pooled embeddings, a predictor maps the image to the continuous embedding of the target caption, which is learned in latent space instead of by autoregressive token generation. Despite this predictive, JEPA-style framing, the training signal is contrastive. VL-JEPA optimizes a bidirectional InfoNCE objective of the same form as CLIP between predicted and target embeddings, which simultaneously aligns each predicted embedding with its matched target and, through the in-batch denominator, pushes it away from the other captions in the batch. Collapse is therefore prevented by negative pairs drawn from the batch rather than by an explicit distributional regularizer, and VL-JEPA inherits the batch-level negative sampling and batch-size dependence of contrastive pretraining; we accordingly group it with the contrastive methods. Notably, \citet{chen2025vljepa} observe that the InfoNCE term could in principle be replaced by a sample-independent anti-collapse regularizer but leave this to future work. 

\subsection{Non-Contrastive Self-Supervised Learning}
In vision-only pretraining, non-contrastive methods have become the dominant paradigm. SimCLR~\citep{chen2020simpleframeworkcontrastivelearning} remains a canonical contrastive baseline, DINO~\citep{caron2021emerging} learns without negatives via self-distillation between student and teacher networks, and I-JEPA~\citep{assran2023self} predicts the representations of masked image regions from visible ones in latent space, avoiding both negatives and pixel-level reconstruction. BYOL~\citep{grill2020bootstrap} prevents collapse without negatives through architectural asymmetry: an online network equipped with a prediction head regresses the representation produced by a momentum-averaged target network, with a stop-gradient applied to the target. This asymmetric predictor-plus-stop-gradient design, which later was simplified by SimSiam~\citep{chen2021exploring}---which shows the momentum encoder is not required---is the mechanism we adapt to the cross-modal setting (Section~\ref{sec:learning-non-contrastive}), and it is complementary to the distributional regularizer of LeJEPA.

LeJEPA~\citep{balestriero2025lejepa} provides a simple and theoretically grounded alternative within the JEPA family by replacing such mechanisms with an explicit distributional regularizer. The central result is that isotropic Gaussian embeddings minimize downstream risk, and SIGReg (Sketched Isotropic Gaussian Regularization) is the objective that enforces this distribution in a scalable way. Rather than matching densities directly in $d$ dimensions, SIGReg projects the embeddings onto a set of random 1D directions $\mathbb{A}$ and applies a characteristic-function-based statistical test (Epps--Pulley) to each projection, yielding an objective with linear time and memory complexity in both batch size and embedding dimension.

Given a batch of $B$ samples with $V$ augmented views per sample ($V_g$ global views and $V_l$ local views), let $z_{n,v} \in \mathbb{R}^d$ denote the embedding of view $v$ of sample $n$. LeJEPA combines SIGReg, applied per-view and averaged across the $V$ views, with a view-prediction loss averaged across the $B$ samples in the batch:
\begin{equation}
\begin{split}
    \mathcal{L}_{\text{LeJEPA}} = {}&\frac{\lambda}{V} \sum_{v=1}^{V} \text{SIGReg}\!\left(\{z_{n,v}\}_{n=1}^{B}\right) \\
    &+ \frac{1-\lambda}{B} \sum_{n=1}^{B} \mathcal{L}_{\text{pred}}^{(V_g)}\!\left(\{z_{n,v}\}_{v=1}^{V}\right),
\end{split}
\end{equation}
with a single trade-off hyperparameter $\lambda \in [0, 1]$. The prediction term draws views closer in latent space while SIGReg keeps the embedding distribution isotropic Gaussian, and together they are sufficient to prevent collapse without stop-gradients, momentum encoders, or teacher-student architectures. We build directly on LeJEPA and reuse its view structure and prediction-plus-SIGReg formulation as the non-contrastive component of our method.

\section{Learning Image-Text Alignment Without Negatives}
\label{sec:learning-non-contrastive}

We investigate whether non-contrastive objectives, of the kind that have replaced contrastive learning in vision-only self-supervised pretraining, can be applied to the vision-language setting. We adopt the LeJEPA framework~\citep{balestriero2025lejepa} as a representative instance: it provides a simple, theoretically grounded non-contrastive objective and serves as our starting point for cross-modal extension. Concretely, given a paired batch $\{(x_i, y_i)\}_{i=1}^B$ with image embeddings $z^v_i = f_\theta(x_i)$ and text embeddings $z^t_i = g_\phi(y_i)$, we want an objective that (i) aligns the two modalities, (ii) regularizes each embedding distribution to be isotropic Gaussian via SIGReg, and (iii) uses no negatives. An overview of the resulting architecture is shown in Figure~\ref{fig:overview}.

\subsection{Direct Alignment Collapses}
A direct adaptation of LeJEPA's view-prediction objective to the cross-modal setting replaces CLIP's contrastive term with a latent-space regression between matched image-text pairs,
\begin{equation}
\begin{split}
    \mathcal{L}_{\text{direct}} = {}&\frac{1}{B} \sum_{i=1}^{B} \| z^v_i - z^t_i \|_2^2 \\
    &+ \lambda_v \, \text{SIGReg}(Z^v) + \lambda_t \, \text{SIGReg}(Z^t),
\end{split}
\end{equation}
trained jointly over both encoders. This symmetric objective is insufficient in practice. Without SIGReg, direct MSE collapses almost completely: the effective rank of $Z^v$ and $Z^t$ drops rapidly during training and the two distributions converge onto a shared low-dimensional subspace. Adding per-modality SIGReg improves the effective rank of both embeddings, but still yields poor zero-shot and linear-probing performance (Appendix~\ref{app:objective-ablations}). Thus, marginal regularization alone does not resolve the degeneracy introduced by symmetric cross-modal regression.

The cause is the symmetry of the regression term. Gradients flow into both encoders from the same target $\| z^v_i - z^t_i \|_2^2$, so the objective encourages both embeddings to meet in a shared space determined by the information common to image and text. Since captions describe images at a coarse level of detail, this shared signal under-specifies fine visual structure. Although SIGReg can pull each marginal distribution toward an isotropic Gaussian and improve rank, the symmetric cross-modal term still couples the two encoders too strongly and does not recover useful cross-modal geometry.

\subsection{Cross-Modal Prediction with Stop-Gradient}
This suggests that stable non-contrastive cross-modal alignment requires an asymmetric objective. Instead of regressing one embedding against the other directly, we introduce a \emph{cross-modal predictor} on each side: a small MLP $h_v : \mathbb{R}^d \to \mathbb{R}^d$ that predicts the text embedding from the image embedding, and a mirror predictor $h_t$ that predicts the image embedding from the text embedding. Crucially, the prediction targets are detached: gradients flow only through the predictor and its own encoder, not back into the encoder producing the target. The cross-modal loss is
\begin{equation}
\begin{split}
    \mathcal{L}_{\text{cross}} = \frac{1}{B} \sum_{i=1}^{B} \Big[ &\| h_v(z^v_i) - \text{sg}(z^t_i) \|_2^2 \\
    &+ \| h_t(z^t_i) - \text{sg}(z^v_i) \|_2^2 \Big],
\end{split}
\end{equation}
where $\text{sg}(\cdot)$ denotes stop-gradient. This asymmetry is the central design choice: with stopped targets, each encoder's gradients come only from its own branch, so SIGReg can shape each modality's embedding distribution without competition from a symmetric cross-modal term. The predictors $h_v, h_t$ absorb the residual asymmetry between image and caption content rather than forcing it into the encoders themselves, similar to the predictor design in SimSiam~\citep{chen2021exploring} and BYOL~\citep{grill2020bootstrap} for vision-only SSL. 

\textbf{Architecture.} 
Both encoders follow the standard CLIP setup: a ViT~\citep{dosovitskiy2020image} image encoder producing a CLS token embedding, and a GPT-2~\citep{radford2019language} text encoder using the final-token hidden state. Each backbone output is passed through a one-layer MLP projection (hidden dimension 2048, GELU, BatchNorm, Dropout) that maps to the shared embedding space of dimension $d$, where we calculate the SIGReg loss. This projection is necessary because both the final ViT layer and the final GPT-2 layer apply a Layer Normalization~\citep{ba2016layernormalization} which prevents the SIGReg anti-collapse objective from being optimized effectively \citep{maes2026leworldmodel}. The cross-modal predictors $h_v, h_t$ are separate MLPs operating on top of these embeddings, so the target embeddings (used with stop-gradient) are exactly the same representations that SIGReg regularizes (see Appendix \ref{app:training-protocol} for details).

\textbf{Objective.} 
The full LeVLJEPA objective combines cross-modal prediction with SIGReg on each modality,
\begin{equation}
\begin{split}
    \mathcal{L}_{\text{LeVLJEPA}} = {}&(1 - \lambda_v - \lambda_t)\, \mathcal{L}_{\text{cross}} \\
    &+ \lambda_v \, \text{SIGReg}(Z^v) + \lambda_t \, \text{SIGReg}(Z^t),
\end{split}
\end{equation}
with a single degree of freedom per modality ($\lambda_v, \lambda_t \in [0,1]$). Unlike contrastive or teacher-student objectives, LeVLJEPA does not introduce negative-pair sampling, temperature scaling, momentum-averaged encoders, or auxiliary teacher networks. The method is trained with a standard forward-backward pass and is compatible with distributed data parallelism.

\subsection{Batch Size Invariance}
Contrastive objectives draw their negative signal from the batch, so their performance depends on the batch size $B$; CLIP-style methods accordingly improve as $B$ grows~\citep{radford2021learning, zhai2023sigmoid}. LeVLJEPA has no batch-level term: cross-modal prediction is computed per sample, and SIGReg acts on the empirical marginal of the embedding distribution, converging to the same population objective regardless of $B$. We confirm this on CC12M across $B \in \{1024, 2048, 4096\}$ at matched compute, where LeVLJEPA's ImageNet zero-shot and linear-probing accuracy stay within noise while CLIP improves with $B$. Non-contrastive cross-modal pretraining is therefore largely insensitive to batch size, removing a constraint that ties contrastive performance to training infrastructure.

\section{Experimental Setup}
\label{sec:experimental-setup}

\paragraph{Two-stage protocol.}
We separate method development from final evaluation to prevent overfitting our design choices to the regime on which we report results. Hyperparameter selection and all objective ablations are conducted on CC12M, whose moderate scale permits sweeping the predictor depth and width, the SIGReg trade-off weights $\lambda_v$ and $\lambda_t$, and the learning rate across a large number of configurations at tractable cost; these studies are reported in Appendix~\ref{app:ablations}. The configuration selected on CC12M is then held fixed and evaluated at the scale of Datacomp-L, where all results in the main paper are obtained. No additional method-specific tuning is performed at the Datacomp-L scale, ensuring that the large-scale results reflect a single transferred configuration rather than a configuration optimized on the evaluation regime itself.

\paragraph{Baselines.}
For the contrastive baselines we use the InfoNCE and SigLIP objectives with the well-tested OpenCLIP~\citep{ilharco_gabriel_2021_5143773} training hyperparameters, which are widely validated for CLIP-style training at this scale, rather than tuning them ourselves. This ensures the contrastive baselines are configured as a practitioner would deploy them and that any gap is not an artifact of under-tuned baselines. All methods share the same ViT-B/16 vision encoder, the same data, the same number of samples seen, and the same evaluation protocols; LeVLJEPA and the contrastive baselines differ only in the pretraining objective.

\paragraph{Scale.}
All Datacomp-L runs are trained for a total of $819$M samples seen. We adopt this samples-seen budget, rather than a fixed number of epochs, as the axis along which all methods are matched, so that total compute is held constant across objectives independently of the effective dataset size each method traverses. Datacomp-L is distributed as a list of image URLs, and a fraction of these are no longer retrievable due to link rot: of the approximately $140$M image-text pairs in the original release, roughly $92$M remained downloadable at the time of our experiments, and all methods are therefore trained on this reduced $\sim\!92$M-sample pool. This bears on cross-paper comparison. Earlier results on Datacomp-L were obtained from substantially larger downloadable fractions, so our absolute numbers are not directly comparable to those reported elsewhere and are expected to be lower. Because link rot affects all methods identically, the comparisons drawn in this paper remain controlled: every encoder is trained on the same surviving pool under the same samples-seen budget.

\paragraph{Training dynamics and ablations.}
Under this protocol LeVLJEPA trains stably from initialization: the effective rank of both $Z^v$ and $Z^t$ rises steadily throughout training until saturation, the SIGReg terms converge to small values consistent with isotropic Gaussian marginals, and the cross-modal loss decreases monotonically. Training is also robust to architectural and optimization choices, with final performance largely insensitive to predictor depth and width and to the trade-off weights $\lambda_v, \lambda_t$ across a wide range (Appendix~\ref{app:ablations}, \ref{app:training-protocol}). Objective ablations on CC12M (Appendix~\ref{app:objective-ablations}) confirm that each component is necessary: direct symmetric MSE collapses, SIGReg alone does not prevent that collapse, predictors without SIGReg are unstable or underperform, and only the full objective is both stable and performant.

\section{Global Representation Quality}
\label{sec:global-readouts}

We first evaluate the encoders under the two standard protocols for vision-language representations: zero-shot transfer and linear probing. Each protocol reduces an image to a single pooled vector prior to scoring---zero-shot transfer compares the pooled image embedding to a text embedding, whereas linear probing classifies the pooled image embedding directly---and both therefore characterize the global representation rather than the per-token features. The two protocols yield distinct conclusions. On zero-shot transfer, the contrastive baselines hold an advantage at the scale of Datacomp-L, consistent with this protocol measuring precisely the image-text alignment they optimize directly. Under linear probing, by contrast, the objectives are on par at both the CC12M and Datacomp-L scales: removing the dependence on cross-modal alignment eliminates the zero-shot gap and leaves the global visual features of all three methods equally discriminative. These protocols thus establish parity in global representation quality while attributing the zero-shot gap to alignment rather than to feature quality. In the following subsections we report detailed results for both protocols across multiple classification benchmarks.

\subsection{Zero-Shot Alignment Without Negatives}
\label{sec:non-contrastive-vl}

Although LeVLJEPA receives no negative-pair signal, cross-modal prediction still induces nontrivial zero-shot alignment. Table~\ref{tab:zeroshot} reports zero-shot accuracy on Datacomp-L across four classification benchmarks. LeVLJEPA produces meaningful predictions across all four, confirming that matched-pair prediction with marginal regularization is sufficient to learn an image-text correspondence usable for classification.

\begin{table}[t]
\centering
\caption{Zero-shot classification accuracy (\%) on Datacomp-L with an identical ViT-B/16 backbone. LeVLJEPA learns nontrivial alignment without negatives but trails the contrastive baselines, which optimize the matched-vs-unmatched objective that zero-shot directly measures. LeVLJEPA uses the text-to-image prediction direction.}
\small
\setlength{\tabcolsep}{4pt}
\begin{tabular}{l cccc}
\toprule
\textbf{Method} & \textbf{ImageNet} & \textbf{Places365} & \textbf{Aircraft} & \textbf{Pets} \\
\midrule
InfoNCE   & 47.32          & \textbf{34.46} & 8.10           & 68.98 \\
SigLIP    & \textbf{50.78} & 33.76          & \textbf{10.62} & \textbf{77.27} \\
\midrule
\rowcolor{tableblue}
LeVLJEPA  & 42.45          & 29.97          & 7.65           & 59.63 \\
\bottomrule
\end{tabular}
\label{tab:zeroshot}
\end{table}

It does, however, trail CLIP and SigLIP. This is expected as zero-shot classification scores matched against unmatched image-text pairs, which corresponds directly to the objective contrastive methods optimize. In contrast, LeVLJEPA is trained only through matched-pair prediction and never sees the discriminative signal this protocol rewards. We therefore do not present zero-shot as a setting where the non-contrastive objective is competitive, but as evidence that alignment emerges without negatives.

The magnitude of this gap is scale-dependent. On CC12M, trained at a batch size of $2048$, all three objectives are nearly on par on zero-shot transfer, with LeVLJEPA exceeding CLIP on the majority of benchmarks (Appendix~\ref{app:cc12m-zeroshot}). The contrastive advantage that is evident at the scale of Datacomp-L is therefore substantially reduced in this regime, consistent with the dependence of contrastive training on large batches: at the moderate batch size used here, the negative signal that drives zero-shot alignment is weaker, and the gap between the objectives narrows accordingly.

\subsection{Global-Feature Linear Probing}
\label{sec:linear-probing}

A linear probe on the frozen CLS token isolates the quality of the global image feature from cross-modal alignment: it discards the text encoder and asks only whether the pooled visual embedding is linearly separable by class. Table~\ref{tab:linear_probing} reports linear-probing accuracy on Datacomp-L. Here the three objectives track closely. LeVLJEPA stays within about $1.5$ points of the strongest baseline on every benchmark and exceeds InfoNCE on Aircraft, with no method dominating across datasets. Removing the dependence on image-text alignment is thus enough to erase the zero-shot gap of Table~\ref{tab:zeroshot}: the global visual feature LeVLJEPA learns is as discriminative as those of CLIP and SigLIP.

\begin{table}[ht]
\centering
\caption{Linear probing accuracy (\%) on frozen CLS features, Datacomp-L with an identical ViT-B/16 backbone. On this global-feature readout the three objectives are on par, in contrast to the zero-shot gap in Table~\ref{tab:zeroshot}.}
\small
\setlength{\tabcolsep}{4pt}
\begin{tabular}{l cccc}
\toprule
\textbf{Method} & \textbf{ImageNet} & \textbf{Places365} & \textbf{Aircraft} & \textbf{Pets} \\
\midrule
InfoNCE   & 65.75          & \textbf{37.11} & 44.10          & \textbf{82.86} \\
SigLIP    & \textbf{66.34} & 36.81          & \textbf{47.46} & 82.64 \\
\midrule
\rowcolor{tableblue}
LeVLJEPA  & 65.42          & 36.07          & 46.38          & 81.28 \\
\bottomrule
\end{tabular}
\label{tab:linear_probing}
\end{table}

This parity is consistent across training scale. On CC12M the three objectives are likewise equivalent under ImageNet linear probing, with LeVLJEPA reaching $31.84\%$ Top-1, against $31.91\%$ for CLIP and $31.83\%$ for SigLIP.

\subsection{Background Robustness}
\label{sec:background-robustness}

IID linear probing measures whether global features are class-discriminative, but not whether they rely on the object itself or on correlated background context. We therefore evaluate background robustness on the ImageNet-9 backgrounds challenge~\citep{xiao2020noise}. A linear classifier is trained on frozen CLS features from the \textit{Original} training split and evaluated without retraining on two controlled variants: \textit{Mixed-Same}, where the foreground is composited onto a background from the same class, and \textit{Mixed-Rand}, where the background is drawn from a randomly selected class. A large accuracy drop under these shifts indicates sensitivity to background cues.

\begin{table}[h]
\centering
\caption{ImageNet-9 background robustness. A linear classifier is trained on frozen CLS features from \textit{Original} images and evaluated without retraining on background-shifted variants. All encoders use the same ViT-B/16 backbone and are pretrained on Datacomp-L.}
\small
\setlength{\tabcolsep}{5pt}
\begin{tabular}{lccc}
\toprule
\textbf{Method} & \textbf{Original} & \textbf{Mixed-Same} & \textbf{Mixed-Rand} \tabularnewline
\midrule
InfoNCE  & 95.98 & 89.41 & 77.31 \tabularnewline
SigLIP   & 96.44 & 89.41 & 78.35 \tabularnewline
\midrule
\rowcolor{tableblue}
LeVLJEPA & \textbf{96.96} & \textbf{91.01} & \textbf{79.75} \tabularnewline
\bottomrule
\end{tabular}
\label{tab:background}
\end{table}

Table~\ref{tab:background} shows that LeVLJEPA attains the highest accuracy in all three settings and exhibits the smallest observed degradation under both background shifts. From Original to Mixed-Same, its accuracy drops by $5.95$ points, compared with $6.57$ for InfoNCE and $7.03$ for SigLIP. Under the stronger Mixed-Rand shift, LeVLJEPA drops by $17.21$ points, versus $18.67$ and $18.09$ points, respectively.

Thus, although the methods are closely matched on standard IID linear probing, LeVLJEPA's global representation is less sensitive to changes in background context. This is consistent with a more object-focused representation. 
\\
\\
Both readouts considered so far reduce the image to a single vector---zero-shot compares it to a caption embedding, linear probing classifies it directly. Modern uses of vision-language encoders, however, do not pool the image at all. In visual instruction tuning, dense prediction, and language-conditioned control, the downstream model attends over the full grid of patch tokens, and its performance is bounded by the semantic quality of those per-token features. In the next section we evaluate encoders on exactly this property---the dense token features their downstream use depends on.

\section{Non-Contrastive Pretraining Yields Stronger Dense Features}
\label{sec:downstream}

 A frozen ViT-B/16 emits a sequence of $196$ patch tokens per image, and it is this sequence, rather than the pooled CLS embedding, that downstream systems consume---a segmentation head classifying each spatial location, or a language model cross-attending to the visual tokens during instruction tuning. The semantic and spatial structure of these tokens is therefore the representational property on which the downstream use of vision-language encoders depends, and it is not constrained by the contrastive image-text objective, which supervises only the pooled embedding.

We assess this property in two stages, in both of which the encoder is held frozen and only a lightweight task head or bridge is trained, so that every difference is attributable to the pretrained representation rather than to downstream adaptation. We first evaluate dense prediction through semantic segmentation, which reads the patch grid with a minimal task head and measures the spatial and semantic structure of the token features directly (Section~\ref{sec:dense-prediction}). We then evaluate the setting these encoders are increasingly deployed in: serving as the frozen visual backbone of a vision-language model, where the patch tokens are consumed by a language model in place of a task head (Section~\ref{sec:vlm-backbone}).

\subsection{Semantic Segmentation}
\label{sec:dense-prediction}

A frozen backbone evaluated through a minimal task head is a direct readout of representation quality: with the encoder held fixed and only a linear head trained on top, accuracy on a dense task is bounded by the spatial and semantic structure already present in the patch tokens. We apply this protocol to semantic segmentation, training a single linear layer on the frozen patch tokens to predict a class label at each location of the $14\times14$ grid, with the same ViT-B/16 backbone across all methods. We report mean intersection-over-union on ADE20K~\citep{zhou2017scene} and COCO-Stuff~\citep{caesar2018coco}.

\begin{table}[h]
\centering
\caption{Linear semantic segmentation (mIoU, \%) on frozen patch tokens, Datacomp-L with an identical ViT-B/16 backbone. A single linear head is trained per method.}
\small
\setlength{\tabcolsep}{6pt}
\begin{tabular}{l cc}
\toprule
\textbf{Method} & \textbf{ADE20K} & \textbf{COCO-Stuff} \\
\midrule
InfoNCE   & 20.90          & 29.02 \\
SigLIP    & 19.24          & 28.88 \\
\midrule
\rowcolor{tableblue}
LeVLJEPA  & \textbf{23.15} & \textbf{31.10} \\
\bottomrule
\end{tabular}
\label{tab:segmentation}
\end{table}

Table~\ref{tab:segmentation} shows a consistent advantage for LeVLJEPA, which exceeds the stronger contrastive baseline by $2.25$ mIoU on ADE20K and by $2.08$ mIoU on COCO-Stuff. This margin is substantially larger than the differences observed on the global readouts of Section~\ref{sec:global-readouts}, and its direction is preserved across two segmentation benchmarks of differing label granularity. Because the task head is a single linear layer, the difference cannot be attributed to head capacity and instead reflects the linear predictability of semantic class from the patch tokens at each spatial location. The contrastive objective supervises only the pooled image-text similarity and shapes the per-token features as a byproduct, which is consistent with the weaker per-location structure observed here.

\subsection{Visual Question Answering with a Frozen Backbone}
\label{sec:vlm-backbone}

A principal contemporary use of a vision-language encoder is as the frozen visual front-end of a vision-language model, in which the patch tokens are projected into a language model's embedding space and consumed by the language model during multimodal instruction following. We evaluate each pretrained encoder in this role under a strict frozen-transfer protocol: both the ViT-B/16 vision encoder and the language model are held frozen, and only a lightweight MLP bridge mapping the patch tokens into the language-model input space is trained on the corresponding train split of each dataset. As neither the encoder nor the language model is updated, downstream accuracy is a function of the pretrained visual features and the bridge alone, isolating the contribution of the encoder in the same manner as the frozen heads of Section~\ref{sec:dense-prediction}. To distinguish properties of the visual features from those of a particular language model, we repeat the full evaluation with two distinct frozen language models, Llama-1B and Qwen-1.5B, and report accuracy on GQA~\citep{hudson2019gqa}, VQAv2~\citep{goyal2017making}, and POPE~\citep{li2023evaluating}. A randomly initialized encoder is included as a floor quantifying the accuracy attainable from the bridge alone. Results are reported in Table~\ref{tab:vlm}; LeVLJEPA attains the highest accuracy in every benchmark--language-model combination, and all pretrained encoders exceed the random floor by a wide margin, confirming that the bridge reads pretrained structure rather than learning the task itself.

\paragraph{Compositional question answering (GQA).}
GQA evaluates compositional reasoning over object attributes, spatial relations, and their conjunctions, and therefore depends on spatially localized visual information rather than a global image descriptor. LeVLJEPA is the strongest backbone under both language models, exceeding the best contrastive baseline by $1.9$ points with Llama-1B and $1.5$ points with Qwen-1.5B. That the advantage on a relational task is realized through a frozen bridge is consistent with the per-location structure observed under semantic segmentation: the spatial information GQA requires is present in the patch tokens prior to any downstream training.

\paragraph{General visual question answering (VQAv2).}
VQAv2 measures open-ended visual question answering across a broad range of object, attribute, and scene queries. The margin in favor of LeVLJEPA is largest on this benchmark, at $2.6$ points with Llama-1B and $4.8$ points with Qwen-1.5B over the strongest contrastive backbone. The contrastive baselines also diverge from each other here---InfoNCE exceeds SigLIP under both language models despite their proximity on the global readouts of Section~\ref{sec:global-readouts}---indicating that the global-feature parity established earlier does not extend to the patch-level features this task consumes.

\paragraph{Object hallucination (POPE).}
POPE evaluates object hallucination through balanced yes/no questions on object presence, on which a backbone that induces a default ``yes'' response inflates recall while hallucinating absent objects; we therefore report accuracy together with F1 and the fraction of ``yes'' responses in Table~\ref{tab:pope}. LeVLJEPA attains the highest accuracy under both language models and the highest F1 under Qwen-1.5B, and produces the most balanced answer distribution, with a yes-rate nearest the calibrated $50\%$, whereas the contrastive backbones and the random floor are skewed toward ``yes.'' Under Llama-1B its F1 falls below the contrastive baselines despite its higher accuracy; this reflects a lower yes-rate, which trades recall on present objects for fewer false positives, rather than weaker hallucination behavior. Across both language models, the LeVLJEPA backbone yields the most accurate and least answer-biased model on this benchmark.

\begin{table}[h]
\centering
\caption{POPE object-hallucination results with frozen encoders and frozen LLMs. Accuracy and F1 are higher-is-better. Bias is the absolute deviation of the yes-rate from the balanced $50\%$, $|\text{yes-rate}-50|$, for which lower is better. Subscripts are standard deviations over three seeds.}
\small
\setlength{\tabcolsep}{5pt}
\begin{tabular}{l ccc}
\toprule
\textbf{Backbone} & Acc & F1 & Bias $\downarrow$ \\
\midrule
\multicolumn{4}{l}{\textit{Llama-1B}} \\
Random   & \mstd{49.6}{1.3} & \mstd{59.3}{1.2} & \mstd{23.8}{6.2} \\
InfoNCE  & \mstd{65.8}{2.4} & \bmstd{69.4}{2.4} & \mstd{11.6}{3.1} \\
SigLIP   & \mstd{62.0}{1.5} & \mstd{68.3}{1.6} & \mstd{19.7}{1.8} \\
\rowcolor{tableblue}
LeVLJEPA & \bmstd{66.9}{2.6} & \mstd{66.6}{1.7} & \bmstd{1.2}{1.6} \\
\midrule
\multicolumn{4}{l}{\textit{Qwen-1.5B}} \\
Random   & \mstd{52.4}{0.7} & \mstd{63.8}{2.4} & \mstd{31.8}{6.8} \\
InfoNCE  & \mstd{71.5}{0.4} & \mstd{73.1}{1.5} & \mstd{5.8}{4.6} \\
SigLIP   & \mstd{70.4}{1.7} & \mstd{72.7}{1.6} & \mstd{8.4}{6.5} \\
\rowcolor{tableblue}
LeVLJEPA & \bmstd{75.0}{1.2} & \bmstd{75.8}{0.9} & \bmstd{3.0}{1.0} \\
\bottomrule
\end{tabular}
\label{tab:pope}
\end{table}

\paragraph{Consistency across language models.}
The ordering of the three pretrained encoders is identical under both language models---LeVLJEPA, then InfoNCE, then SigLIP. As Llama-1B and Qwen-1.5B differ in tokenizer, pretraining corpus, and architecture, an ordering preserved across the substitution reflects a property of the visual features rather than an interaction with a specific language model. This comparison further dissociates backbone quality from zero-shot alignment: SigLIP is the strongest zero-shot model in Table~\ref{tab:zeroshot} yet the weakest of the three backbones, whereas LeVLJEPA is the weakest zero-shot model yet the strongest backbone. On these encoders, zero-shot accuracy is thus not merely uninformative of backbone quality but inversely related to it, providing direct evidence for the central claim of this work: the representational property that yields a strong vision-language classifier is distinct from the one that yields a strong visual feature encoder.

\begin{table*}[t]
\centering
\caption{Frozen VLM-backbone accuracy (\%) on GQA, VQAv2, and POPE, with two frozen language models. Both the ViT-B/16 vision encoder and the LLM are frozen; only an MLP bridge is trained. All encoders are pretrained on Datacomp-L. Subscripts are standard deviations over three bridge-training seeds. LeVLJEPA is the strongest backbone in every column.}
\setlength{\tabcolsep}{8pt}
\begin{tabular}{l ccc ccc}
\toprule
& \multicolumn{3}{c}{\textbf{Llama-1B}} & \multicolumn{3}{c}{\textbf{Qwen-1.5B}} \\
\cmidrule(lr){2-4} \cmidrule(lr){5-7}
\textbf{Backbone} & GQA & VQAv2 & POPE & GQA & VQAv2 & POPE \\
\midrule
Random      & \mstd{36.4}{0.6} & \mstd{43.2}{0.1} & \mstd{49.6}{1.3} & \mstd{37.0}{0.7} & \mstd{43.6}{0.4} & \mstd{52.4}{0.7} \\
InfoNCE     & \mstd{42.7}{0.4} & \mstd{51.6}{0.4} & \mstd{65.8}{2.4} & \mstd{42.2}{0.9} & \mstd{49.4}{0.4} & \mstd{71.5}{0.4} \\
SigLIP      & \mstd{42.4}{0.3} & \mstd{49.2}{0.2} & \mstd{62.0}{1.5} & \mstd{41.6}{0.3} & \mstd{47.7}{0.6} & \mstd{70.4}{1.7} \\
\midrule
\rowcolor{tableblue}
LeVLJEPA    & \bmstd{44.6}{0.2} & \bmstd{54.2}{1.4} & \bmstd{66.9}{2.6} & \bmstd{43.7}{0.2} & \bmstd{54.1}{0.7} & \bmstd{75.0}{1.2} \\
\bottomrule
\end{tabular}
\label{tab:vlm}
\end{table*}

These results extend the dense-prediction findings to the deployment setting. The advantage observed as stronger per-location semantic structure under semantic segmentation manifests, when the encoder serves as a frozen visual front-end, as a more accurate and better-calibrated vision-language model, across two language models and three benchmarks and without updating either the encoder or the language model. Non-contrastive vision-language pretraining thus yields the dense semantic features on which downstream language-model use depends.

\section{Discussion}

LeVLJEPA demonstrates that end-to-end vision-language pretraining can be performed without contrastive negatives. Using only matched-pair cross-modal prediction with stop-gradient targets and per-modality distributional regularization and no negatives, temperature, momentum encoder, or teacher-student schedule, it trains stably at the scale of Datacomp-L and learns representations that match contrastive pretraining on global-feature readouts. Negative-pair discrimination can therefore bee seen as only one route to useful vision-language representations rather than a precondition for them.

The central finding of this work is that non-contrastive, JEPA-style vision-language pretraining yields a vision encoder with markedly stronger dense semantic features than contrastive pretraining. As the frozen backbone of a vision-language model, LeVLJEPA is the strongest of the evaluated encoders across GQA, VQAv2, and POPE under two distinct language models, and it outperforms both contrastive baselines on semantic segmentation. This advantage is specific to evaluations that consume the full patch-token sequence: under zero-shot transfer and linear probing, both of which reduce an image to a single pooled vector, the three objectives are difficult to separate, with the contrastive baselines retaining an advantage only on the zero-shot alignment they optimize directly. The standard evaluation protocols therefore do not measure the property on which these encoders are now deployed, and the divergence between objectives is visible only at the token level.

These results motivate a vision-centric formulation of vision-language pretraining. Rather than optimizing the global image-text alignment rewarded by classification metrics, the objective is to extract as much usable structure as possible from both modalities and to produce a vision encoder whose dense per-token features serve the systems that now consume them---visual instruction tuning, dense prediction, and language-conditioned control, none of which read the pooled embedding. The non-contrastive predictive objective is suited to this formulation: by supervising each modality through prediction and regularizing each marginal toward isotropy, rather than collapsing the image to a single vector matched against a caption, it preserves stronger semantic structure in the patch tokens that downstream models attend over. The background-robustness results are consistent with this account: even under a global readout, LeVLJEPA's representation is less sensitive to background substitution than the contrastive baselines, indicating a more object-centric representation of the kind that dense, language-grounded use depends on.

Several questions remain open. The contrastive objectives retain a stronger mechanism for zero-shot image-text alignment, and we do not close this gap; whether the dense-feature advantage of non-contrastive pretraining can be combined with competitive alignment within a single objective is a natural direction for future work. Our experiments are conducted with a ViT-B/16 backbone, and while LeVLJEPA trains stably and remains competitive at the scale of Datacomp-L, establishing that these advantages persist at larger model and data scales remains important to verify. More broadly, these results indicate that as vision-language encoders are increasingly deployed as the visual front-end of language models, the objectives and evaluation protocols used to train and compare them should be reconsidered around the dense token features on which these systems rely.

\paragraph{Acknowledgments}

We gratefully acknowledge support from the hessian.AI Service Center (funded by the Federal Ministry of Research, Technology and Space, BMFTR, grant no. 16IS22091) and the hessian.AI Innovation Lab (funded by the Hessian Ministry for Digital Strategy and Innovation, grant no. S-DIW04/0013/003).

This work was co-funded by the European Union (ERC, TAIPO, 101088594 to F.B.) grant. Views and opinions expressed are those of the authors only and do not necessarily reflect those of the European Union or ERC. Neither the European Union nor the granting authority can be held responsible for them.

\newpage

\bibliography{main}

\clearpage
\appendix

\section{Training Protocol and Hyperparameter Tuning}
\label{app:training-protocol}

All methods are trained under a matched compute budget and a shared optimization schedule, so that reported differences reflect the pretraining objective rather than differences in training configuration. Each model uses the same ViT-B/16 backbone, the same global batch size, the same warmup, and the same cosine learning-rate decay schedule, and is evaluated under an identical protocol. Hyperparameter selection is performed on CC12M and the resulting configuration is transferred unchanged to the Datacomp-L runs on which the main results are reported. We are using stable-pretraining \citep{balestriero2025stable} throughout all experiments.

\paragraph{Optimization and hardware.}
Each model is trained on $16$ GPUs with a per-rank batch size of $256$, for a global batch size of $4096$. Training uses a $1{,}000$-step warmup followed by cosine decay. Each run requires approximately $256$ H100 GPU-hours.

\paragraph{Learning-rate tuning.}
The learning rate is tuned independently for each method over a small grid on CC12M, and the best-performing configuration under the shared schedule is retained. This allows each objective to operate at an appropriate step size while holding the data, backbone, number of updates, warmup, and evaluation protocol fixed. At the Datacomp-L scale, the contrastive baselines instead use the InfoNCE and SigLIP objectives with the established OpenCLIP~\citep{ilharco_gabriel_2021_5143773} hyperparameters, which are well validated for CLIP-style training at this scale, rather than a learning rate tuned by us; this ensures the contrastive baselines are configured as a practitioner would deploy them and that any observed gap is not an artifact of under-tuned baselines.

\paragraph{LeVLJEPA configuration.}
LeVLJEPA uses a learning rate of $4\times10^{-2}$, which we found stable across the batch sizes considered. The vision and text SIGReg weights are set to $\lambda_v=\lambda_t=0.01$. The cross-modal predictor is a depth-$4$ MLP of width $2048$ with BatchNorm and GELU, and applies 10\% dropout to limit overfitting at scale. SIGReg is computedwith cross-device gathering similar to the contrastive baselines.

\paragraph{}
This protocol compares objectives under matched compute rather than maximizing absolute performance through method-specific scaling recipes. All reported differences should accordingly be interpreted as differences under the same data, backbone, number of updates, warmup, and evaluation setup, with the learning rate tuned per method.

\section{Datasets}
\label{app:datasets}

\paragraph{Pretraining.}
We pretrain on two image-text datasets. CC12M~\citep{changpinyo2021conceptual} contains approximately $12$M image-caption pairs and is used for hyperparameter selection, ablations, and small-scale validation. Datacomp-L~\citep{gadre2023datacomp} is used for the main results; of its $\sim\!140$M originally released pairs, $\sim\!92$M remained downloadable at the time of our experiments due to link rot (Appendix~\ref{app:training-protocol}), and all models are trained on this surviving pool.

\paragraph{Classification (zero-shot and linear probing).}
We evaluate on ImageNet~\citep{deng2009imagenet} ($1000$ classes), Places365~\citep{zhou2017places} (scene recognition, $365$ classes), FGVC-Aircraft~\citep{maji13fine-grained} (fine-grained recognition, $100$ classes), and Oxford-IIIT Pets~\citep{parkhi2012cats} ($37$ classes).

\paragraph{Semantic segmentation.}
ADE20K~\citep{zhou2017scene} contains $20{,}210$ training and $2{,}000$ validation images across $150$ semantic classes. COCO-Stuff~\citep{caesar2018coco} provides dense annotations over $171$ classes. Both are evaluated with a linear head on frozen patch tokens at the $14\times14$ grid resolution.

\paragraph{Background robustness.}
ImageNet-9~\citep{xiao2020noise} is a $9$-superclass subset of ImageNet with controlled foreground/background recompositions. We use the \textit{Original}, \textit{Mixed-Same}, and \textit{Mixed-Rand} variants, training a linear classifier on \textit{Original} and evaluating on all three.

\paragraph{Vision-language model evaluation.}
We evaluate the frozen-backbone VLMs on GQA~\citep{hudson2019gqa} (compositional question answering), VQAv2~\citep{goyal2017making} (general visual question answering), and POPE~\citep{li2023evaluating} (object hallucination, balanced yes/no presence questions).

\section{Evaluation Protocol}
\label{app:evaluation-protocol}

For linear probing we evaluate frozen ViT-B/16 vision features without the projection head. For each image we extract the CLS token from the frozen vision encoder, yielding a $768$-dimensional feature vector, and apply $\ell_2$ normalization. Images are resized so that the shortest side is $256$, center-cropped to $224\times224$, and normalized with the ImageNet mean and standard deviation.

\paragraph{Linear probing.}
The linear probe is a single linear layer from the frozen feature dimension to the number of classes, without bias. Prior to fitting, features are standardized per dimension using the training-set mean and standard deviation, with standard deviations clamped at $10^{-8}$, and the same statistics are applied to the test set. The probe is trained for $50$ epochs with AdamW at learning rate $0.1$, weight decay $10^{-4}$, and batch size $4096$, with the training set reshuffled each epoch.

\paragraph{Zero-shot evaluation.}
For zero-shot classification we use the frozen image and text encoders without fitting any classifier. Unless stated otherwise, class names are embedded with the single prompt template
\[
\texttt{a photo of a \{class\}}.
\]
Image embeddings are compared against text embeddings by cosine similarity, and the nearest text class is taken as the prediction. We apply this single-template protocol to all methods to keep the evaluation controlled and to avoid confounding objective-level differences with prompt engineering, prompt ensembling, or dataset-specific calibration. These choices lower absolute zero-shot accuracy relative to fully tuned CLIP-style evaluations but yield a cleaner comparison between objectives. We therefore interpret the zero-shot results of Table~\ref{tab:zeroshot} as a controlled comparison of training objectives rather than as an attempt to maximize absolute accuracy.

\section{Objective Ablations}
\label{app:objective-ablations}

We ablate the components of LeVLJEPA to test which parts are necessary for stable non-contrastive vision-language pretraining. We compare four variants: direct latent MSE between matched image and text embeddings, direct MSE with per-modality SIGReg, cross-modal predictors with stop-gradient targets but without SIGReg, and the full LeVLJEPA objective. All ablations are run for $20{,}000$ training steps under the same reduced experimental setting; absolute numbers are therefore not directly comparable to the main results.

\begin{table*}[t]
\centering
\caption{Objective ablation for LeVLJEPA after $20{,}000$ training steps. We report whether SIGReg is used, effective rank of the text and vision embeddings, ImageNet zero-shot Top-1, and ImageNet linear probing Top-1. Direct symmetric alignment collapses. Adding SIGReg improves rank but remains insufficient without predictor/stop-gradient asymmetry. Predictors with stop-gradient improve zero-shot alignment but underperform without marginal regularization. The full LeVLJEPA objective is required for stable and performant non-contrastive vision-language pretraining.}
\small
\begin{tabular}{l c cc cc}
\toprule
\textbf{Alignment Objective} & \textbf{SIGReg} & \textbf{Eff.\ rank Text} & \textbf{Eff.\ rank Vision} & \textbf{ImageNet ZS} & \textbf{ImageNet LP} \\
\midrule
Direct MSE & No  & 25  & 3   & 0.52  & 0.18 \\
Direct MSE & Yes & 63  & 170 & 1.82  & 2.56 \\
Predictor + stop-gradient & No  & 27  & 161 & 9.40  & 14.36 \\
\midrule
\rowcolor{tableblue}
Predictor + stop-gradient & Yes & \textbf{358} & \textbf{477} & \textbf{25.24} & \textbf{27.51} \\
\bottomrule
\end{tabular}
\vspace{6pt}
\label{tab:objective_ablations}
\end{table*}

The results in Table~\ref{tab:objective_ablations} show that symmetric cross-modal regression is not sufficient. Direct MSE collapses almost completely, especially on the vision side, yielding effective rank $3$ and near-random ImageNet performance. Adding SIGReg increases the effective rank, but does not recover useful image-text alignment or transferable vision features, indicating that marginal regularization alone cannot overcome the degeneracy induced by symmetric regression. Introducing predictors and stop-gradient targets improves zero-shot alignment substantially, but remains unstable and produces weak linear probing without SIGReg. The full LeVLJEPA objective combines both ingredients: the predictor/stop-gradient asymmetry stabilizes cross-modal alignment, while SIGReg preserves high-rank modality-specific representations. This combination yields the highest effective ranks and the strongest zero-shot and linear probing performance among the ablated variants.

\section{CC12M Global Representation Results}
\label{app:cc12m-zeroshot}

This appendix reports the full CC12M results summarized in Section~\ref{sec:global-readouts}. All models use an identical ViT-B/16 backbone and are trained on CC12M at a batch size of $2048$ under the protocol of Appendix~\ref{app:training-protocol}.

\paragraph{Zero-shot transfer.}
Table~\ref{tab:cc12m_zeroshot} reports zero-shot accuracy across five classification benchmarks. At this scale the three objectives are nearly on par: LeVLJEPA exceeds CLIP on STL-10, ImageNet Top-1, CIFAR-100 Top-1, Places365, and SUN397, and matches SigLIP on SUN397 Top-1, while SigLIP retains an advantage on ImageNet and CIFAR-100. LeVLJEPA's largest margins occur on scene classification, where it improves over CLIP by $4.6$ Top-1 and $6.3$ Top-5 on Places365 and by $2.8$ Top-1 and $4.6$ Top-5 on SUN397. The contrastive advantage observed on zero-shot transfer at the scale of Datacomp-L (Table~\ref{tab:zeroshot}) is thus substantially reduced in this regime, consistent with the batch-size dependence of contrastive training.

\begin{table*}[h]
\centering
\caption{Zero-shot classification accuracy (\%) on CC12M at batch size $2048$ with an identical ViT-B/16 backbone. At this scale the three objectives are nearly on par, with LeVLJEPA strongest on STL-10 and scene classification.}
\small
\begin{tabular}{l c cc cc cc cc}
\toprule
\textbf{Method} & \textbf{STL-10} & \multicolumn{2}{c}{\textbf{ImageNet}} & \multicolumn{2}{c}{\textbf{CIFAR-100}} & \multicolumn{2}{c}{\textbf{Places365}} & \multicolumn{2}{c}{\textbf{SUN397}} \\
\cmidrule(lr){2-2}\cmidrule(lr){3-4}\cmidrule(lr){5-6}\cmidrule(lr){7-8}\cmidrule(lr){9-10}
 & Top-1 & Top-1 & Top-5 & Top-1 & Top-5 & Top-1 & Top-5 & Top-1 & Top-5 \\
\midrule
InfoNCE        & 68.05          & 12.39          & 28.55          & 9.77           & 30.74          & 15.68          & 38.78          & 22.43          & 49.36 \\
SigLIP         & 68.33          & \textbf{14.05} & \textbf{31.28} & \textbf{13.77} & \textbf{36.06} & 18.40          & 43.45          & \textbf{25.27} & \textbf{54.78} \\
\midrule
\rowcolor{tableblue}
LeVLJEPA       & \textbf{74.16} & 12.84          & 30.23          & 11.03          & 29.70          & \textbf{20.24} & \textbf{45.04} & 25.27          & 53.98 \\
\bottomrule
\end{tabular}
\label{tab:cc12m_zeroshot}
\end{table*}

\paragraph{Linear probing.}
Under linear probing of the frozen CLS feature on ImageNet, the three objectives are indistinguishable: LeVLJEPA reaches $31.84\%$ Top-1, against $31.91\%$ for CLIP and $31.83\%$ for SigLIP. This matches the parity observed under linear probing at the scale of Datacomp-L (Table~\ref{tab:linear_probing}), indicating that the equivalence of the global visual features across objectives is not specific to scale.

\section{Hyperparameter Sensitivity}
\label{app:ablations}

We study LeVLJEPA's sensitivity to two architectural and optimization choices: the depth of the cross-modal predictor and the SIGReg trade-off weight $\lambda_v$ and $\lambda_t$. All experiments use the configuration described in the main paper except for the ablated dimension; numbers are ImageNet zero-shot and linear probing Top-1 at batch size $2048$ on CC12M.

\paragraph{Predictor depth.}
\begin{figure}[h]
  \centering
  \includegraphics[width=\linewidth]{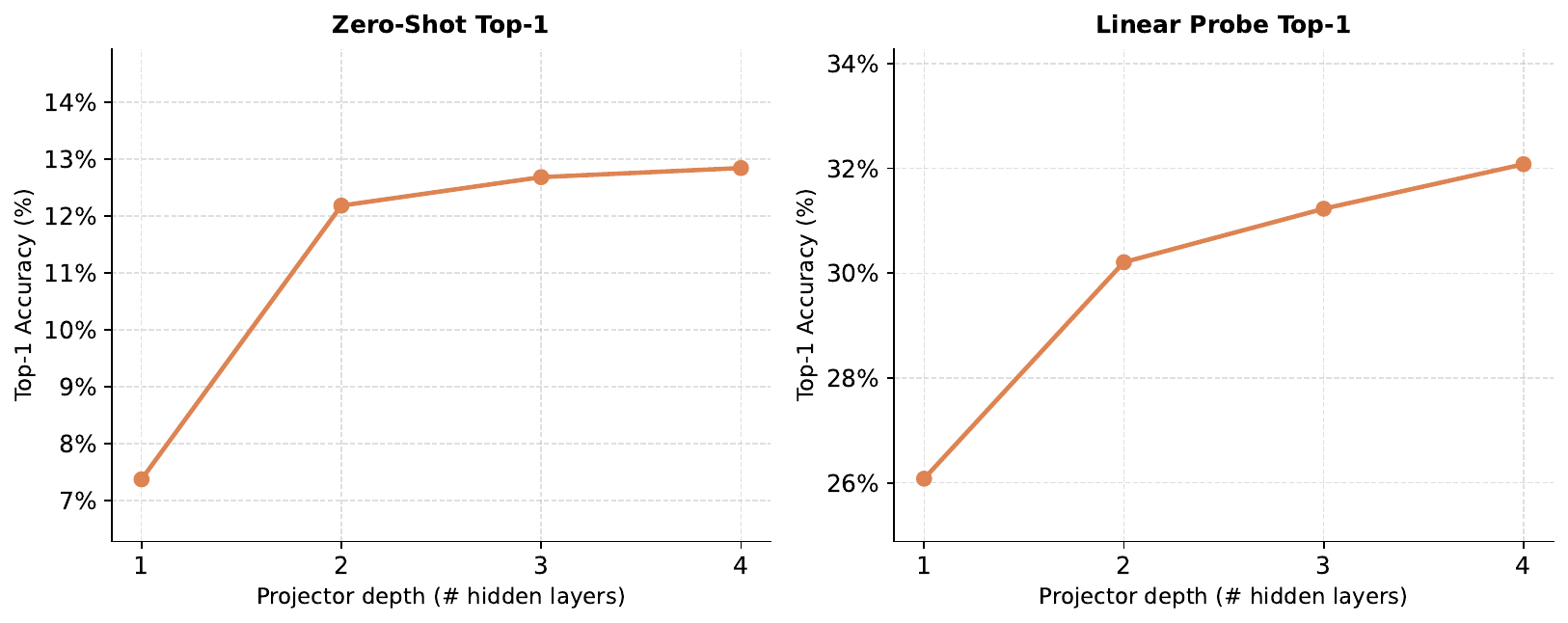}
  \caption{\textbf{Sensitivity to predictor depth.} Zero-shot (left) and linear probing (right) Top-1 accuracy as a function of the number of hidden layers in the cross-modal predictor.}
  \label{fig:ablation_depth}
\end{figure}

Figure~\ref{fig:ablation_depth} sweeps the cross-modal predictor's depth from $1$ to $4$ hidden layers. A single hidden layer is insufficient. Zero-shot drops to $7.4\%$ and linear probing to $26.1\%$, both well below the configurations with more capacity. Beyond two layers, performance is stable and improves monotonically with depth, gaining $0.7$ points on zero-shot and $1.9$ points on linear probing between depths $2$ and $4$. The default configuration of $4$ hidden layers used throughout the paper sits within the stable region.

\paragraph{SIGReg weight.}
\begin{figure}[h]
  \centering
  \includegraphics[width=\linewidth]{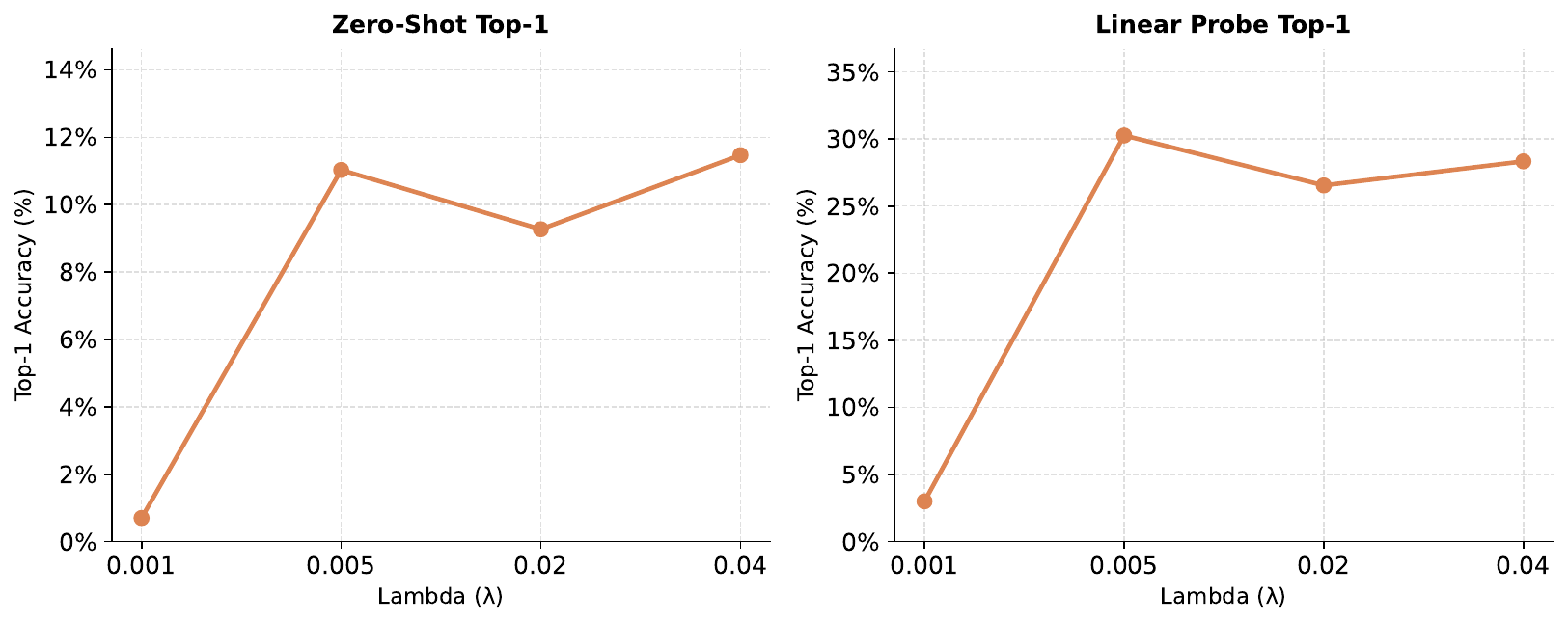}
  \caption{\textbf{Sensitivity to the SIGReg trade-off weight $\lambda$.} Zero-shot (left) and linear probing (right) Top-1 accuracy as a function of $\lambda$ on a log scale.}
  \label{fig:ablation_lambda}
\end{figure}

Figure~\ref{fig:ablation_lambda} sweeps the SIGReg weights $\lambda_v$ and $\lambda_t$ (tested symmetrically and denoted as $\lambda$) across two orders of magnitude. At $\lambda = 0.001$, SIGReg is too weak to enforce isotropy and training degenerates: zero-shot drops below $1\%$ and linear probing to $3\%$. For $\lambda \in [0.005, 0.04]$, performance is stable, with probing accuracy varying within a $4$-point range and zero-shot within a $2$-point range. LeVLJEPA does not require precise tuning of $\lambda$ as long as SIGReg is given enough weight to actively shape the embedding distribution.

Together, these ablations indicate that LeVLJEPA is robust to architectural and optimization choices once minimum thresholds (two predictor layers and a sufficiently strong SIGReg term) are satisfied.

\section{Scaling}
\subsection{Scaling Model Size}
\label{app:model-scaling}

\begin{figure}[h]
  \centering
  \includegraphics[width=\linewidth]{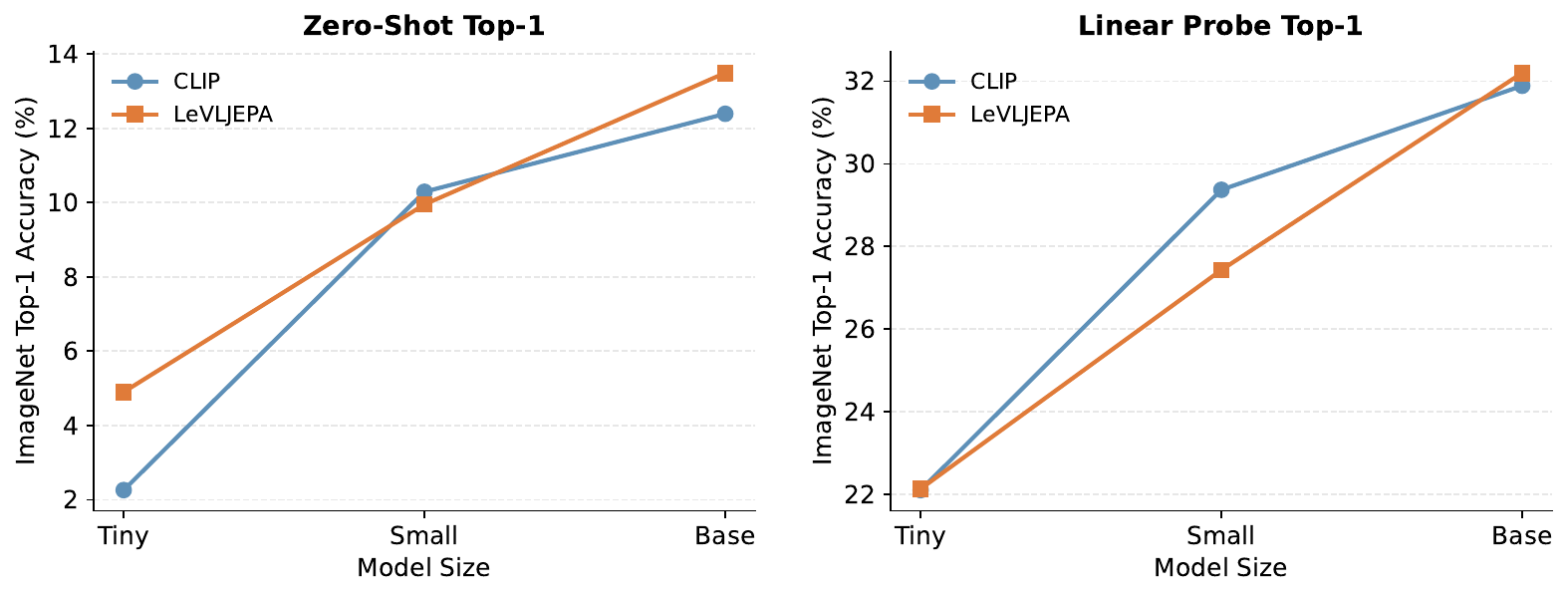}
  \caption{\textbf{Model size scaling on CC12M.} LeVLJEPA vs.\ CLIP on ImageNet zero-shot (left) and linear probing (right) across three ViT backbone sizes: Tiny, Small, and Base. Both methods improve with model size, and LeVLJEPA remains competitive with CLIP across the full range.}
  \label{fig:model_scaling}
\end{figure}

Figure~\ref{fig:model_scaling} reports ImageNet performance across three ViT backbone sizes (Tiny, Small, Base) under the same update budget. Both methods benefit from increased model capacity. On zero-shot classification, LeVLJEPA outperforms CLIP at Tiny ($4.90\%$ vs.\ $2.26\%$), is within $0.34$ points at Small ($9.95\%$ vs.\ $10.29\%$), and slightly exceeds CLIP at Base ($13.49\%$ vs.\ $12.39\%$). Linear probing shows the same competitive pattern: the two methods are essentially tied at Tiny ($22.13\%$ vs.\ $22.10\%$), CLIP is ahead at Small ($29.37\%$ vs.\ $27.43\%$), and LeVLJEPA slightly exceeds CLIP at Base ($32.20\%$ vs.\ $31.89\%$). These results suggest that the non-contrastive formulation scales with model capacity and is not restricted to a single backbone size.

\subsection{Data Efficiency}
\label{app:data-scaling}

\begin{figure}[h]
  \centering
  \includegraphics[width=\linewidth]{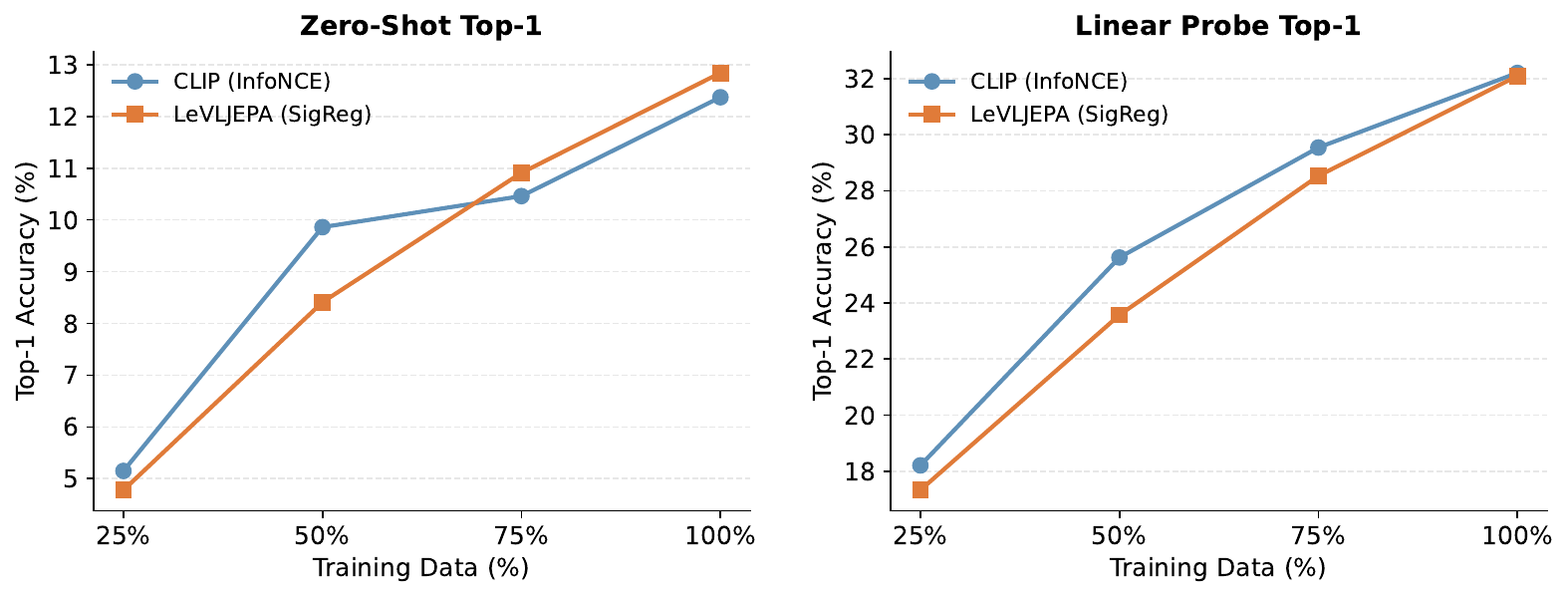}
  \caption{\textbf{Data efficiency on CC12M.} LeVLJEPA vs.\ CLIP (InfoNCE) on ImageNet zero-shot (left) and linear probing (right) when trained on $25\%$, $50\%$, $75\%$, and $100\%$ of CC12M under the same update budget. Both methods improve steadily with more data and converge to nearly identical linear probing accuracy at the full dataset.}
  \label{fig:data_scaling}
\end{figure}

Figure~\ref{fig:data_scaling} reports ImageNet zero-shot and linear probing accuracy when both methods are trained on subsets of CC12M under the same update budget. Both methods scale smoothly with dataset size. At smaller data fractions, CLIP holds a modest advantage: at $25\%$ of CC12M, CLIP reaches $5.15\%$ zero-shot and $18.21\%$ linear probing Top-1, compared to $4.78\%$ and $17.33\%$ for LeVLJEPA; at $50\%$, CLIP reaches $9.86\%$ zero-shot and $25.62\%$ linear probing, compared to $8.41\%$ and $23.57\%$ for LeVLJEPA. The gap narrows as data increases. At $75\%$, LeVLJEPA slightly exceeds CLIP on zero-shot accuracy ($10.91\%$ vs.\ $10.46\%$), while CLIP remains ahead on linear probing ($29.54\%$ vs.\ $28.53\%$). At the full dataset, the methods are nearly tied on linear probing ($32.20\%$ for CLIP vs.\ $32.09\%$ for LeVLJEPA), while LeVLJEPA slightly exceeds CLIP in zero-shot accuracy ($12.84\%$ vs.\ $12.37\%$). These results indicate that LeVLJEPA scales competitively with data under the matched training regime.

\newpage

\end{document}